\documentclass[runningheads]{llncs}
\usepackage[T1]{fontenc}
\usepackage{graphicx}
\usepackage{amsmath}
\usepackage{amsfonts}
\usepackage{color} 

\usepackage{multicol}
\usepackage{fancyhdr}
\fancypagestyle{firstpage}{%
    \chead{This paper has been accepted for publication at the 2024 International Conference on Medical Image Computing and Computer Assisted Intervention.}
    }

\begin{document}
\title{A Patient-Specific Framework for Autonomous Spinal Fixation via a Steerable Drilling Robot}

%
\titlerunning{A Patient-Specific Framework for Autonomous
Spinal Fixation }
%
\author{Susheela Sharma$^1$, Sarah Go$^1$, Zeynep Yakay$^1$, Yash Kulkarni$^1$, Siddhartha Kapuria$^1$, Jordan P. Amadio$^2$, Mohsen Khadem$^3$, Nassir Navab$^4$, and Farshid Alambeigi$^1$}
\authorrunning{S. Sharma et al.}
%
\institute{$^1$Texas Robotics, The University of Texas at Austin\\
$^2$The University of Texas Dell Medical School\\
$^3$The School of Informatics, The University of Edinburgh\\
$^4$Chair for Computer Aided Medical Procedures and Augmented Reality, Technical University of Munich, Garching, Germany\\
\vspace{20pt}
\textbf{THIS PAPER HAS BEEN ACCEPTED FOR PUBLICATION AT THE 2024 INTERNATIONAL CONFERENCE ON MEDICAL IMAGE COMPUTING AND COMPUTER ASSISTED INTERVENTION.}\\}
\maketitle              
\begin{abstract}
In this paper, with the goal of enhancing the minimally invasive spinal fixation procedure in osteoporotic patients,  we propose a first-of-its-kind image-guided robotic framework  for performing an autonomous and patient-specific procedure using a unique concentric tube steerable drilling robot (CT-SDR). Particularly, leveraging a CT-SDR, we introduce the concept of J-shape drilling  based on a pre-operative  trajectory planned in CT scan of a patient followed by appropriate calibration,  registration, and navigation steps to safely execute this trajectory in real-time using our unique robotic setup. 
To thoroughly evaluate the performance of our framework, we  performed several  experiments on two different vertebral phantoms  designed based on CT scan of real patients.

\end{abstract}

\section{Introduction}

Pedicle Screw Fixation (PSF)  is a minimally invasive image-guided surgical procedure that is widely used in spine surgery to stabilize and support the spinal column. The procedure starts with utilizing a drilling instrument to create tunnels with linear trajectories through the pedicle of the vertebra and then implantation  of rigid pedicle screws into the drilled tunnels \cite{screw_place,Esfandiari2018ADL}. This technique relies heavily on intraoperative fluoroscopic images taken to guide the surgeon during the drilling and screw  placement within the vertebra to ensure safety and accuracy of this procedure. 
Despite the advantages of this procedure, rigidity  of the utilized surgical instruments and pedicle screws together with complex anatomy of vertebra impose several challenges for the clinicians, including: (C1) lack of instrument steerability and limited access to the vertebral body  hindering clinicians to achieve optimal angles for drilling trajectories and screw placement. This problem becomes more critical in patients experiencing osteoporosis as their low bone density increases the risk of screw pullout and fixation failure \cite{wittenberg1991importance,okuyama1993stability,weiser2017insufficient}; (C2) complexity and  steep learning curve associated with this procedure demanding  surgeons with a high level of skill and expertise; and (C3) compromised safety and accuracy due to the lack of instrument maneuverability around critical anatomical structures leading to screw misplacement and failure.

To partially address these limitations (mainly C1 and C2), various navigation and  robotic technologies have been introduced \cite{Li2023RoboticSA}. Surgical navigation systems leverage preoperative (e.g., CT images) and/or intraoperative imaging (e.g., fluoroscopy) to create a detailed 3D model of the patient's vertebra \cite{CT_MRI_registration,stent_recovery}. Utilizing optical tracking systems, surgeons are then able to track their instruments in real-time relative to this model  to improve the accuracy of drilling and screw placement and, therefore, reduce the risk of  screw misplacement and potential nerve damage \cite{Li2023RoboticSA,ElmiTerander2018PedicleSP}. Despite the benefits of this approach, the procedure still heavily relies on the clinician's skills to perform the procedure. To mitigate this issue,  commercially available surgical robotic systems (e.g., Mazor X (Medtronic, Dublin, Ireland) and the  Rosa One (Zimmer Biomet, Warsaw, IN, USA)), offer even greater precision by assisting or automating the drilling procedure and placement of screws. In these systems,  the patient's vertebra is first registered to the obtained CT scan of the patient. Next,  navigation and  robotic system assist clinicians to guide the instruments to perform the drilling process \cite{ElmiTerander2018PedicleSP}. Despite the benefits of these systems, due to the use of rigid instrumentation and pedicle screws, these advanced technologies still suffer from the same aforementioned challenges observed in the manual procedures, i.e., lack of access to high bone density regions. Moreover, due to the limited available workspace constrained by rigidity of instruments and vertebra's anatomy,  no real surgical planning is performed in these systems.
 
\begin{figure}[t!] 
	\centering 
	\includegraphics[width=0.95\linewidth]{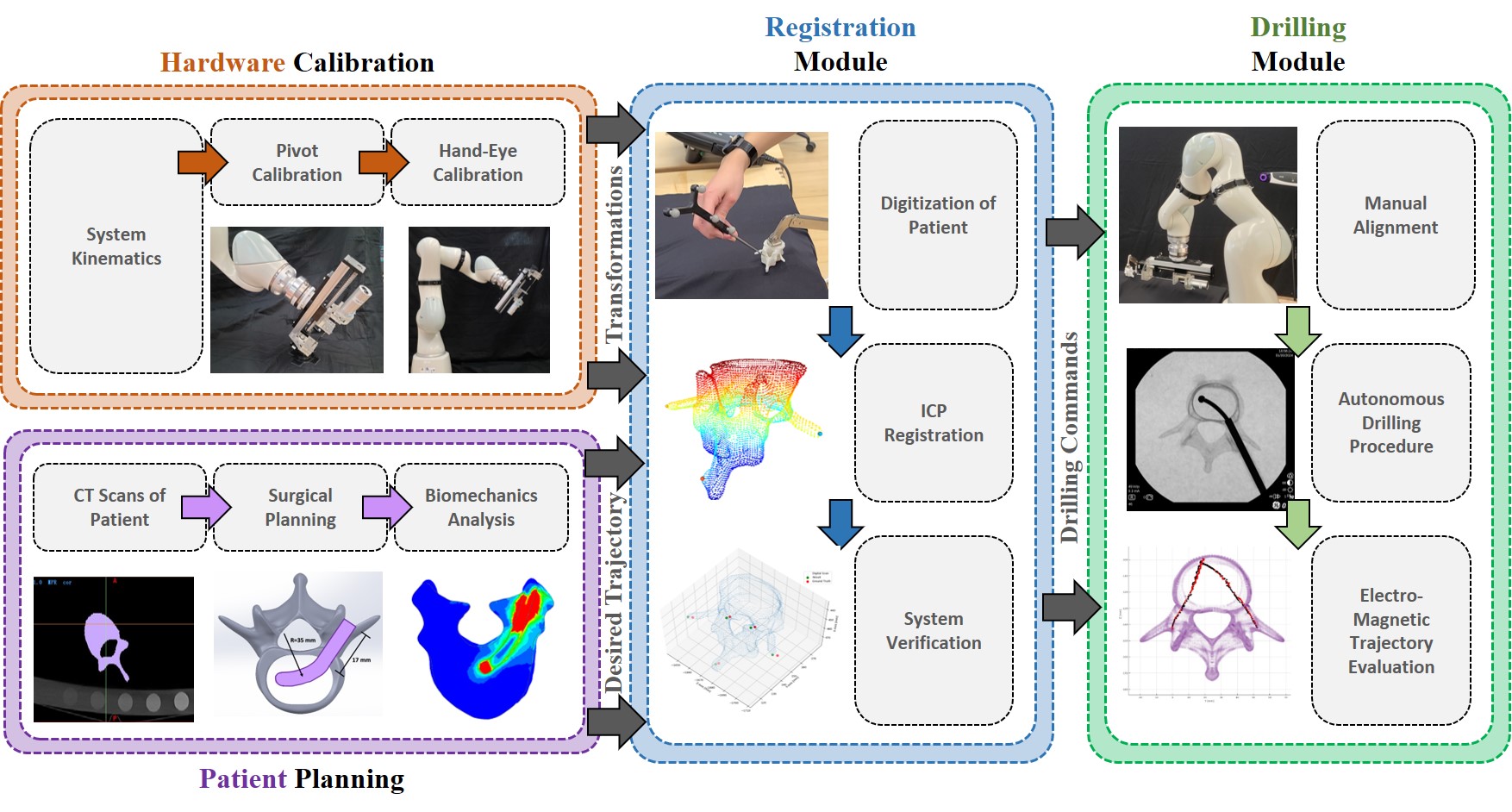}
	\caption{The overall proposed framework including a patient planning module and calibration and registration methods.}
	\label{fig:framework}
\end{figure}
Several researchers have introduced steerable drilling robots (SDRs) to access difficult-to-reach  areas within anatomies with complex geometries (e.g., spine and pelvis)  for placing  implants in high-density bone regions (addressing C1). This includes Articulated Hinged Drilling Tools \cite{wang2021design,9732206}, 
Tendon-Driven SDRs \cite{alambeigi2020steerable,alambeigi2019use,alambeigi2017curved}, and
Concentric Tube Steerable Drilling Robot (CT-SDR) \cite{Sharma_tbme_2022,Sharma_icra,Sharma_ismr}. While these innovations, particularly the CT-SDRs, represent significant progress in addressing surgical accessibility and precision, they also introduce new challenges in their ease of use  (i.e., C2),  change of surgical workflow, and operation time \cite{alambeigi2017curved,Sharma_tbme_2022}.  These limitations necessitate  reevaluation of existing planning, navigation, and control techniques to allow drilling patient-specific complex nonlinear paths for not compromising patient safety and ideal implant positioning.

In this paper, to holistically address the  aforementioned challenges of the existing manual and robot-assisted PSF procedures, we propose a first-of-its-kind image-guided robotic framework (shown in Fig. \ref{fig:framework})  for performing an autonomous and patient-specific procedure using a CT-SDR for normal and osteoporotic patients. Particularly, for the first time, we introduce the concept of autonomous minimally invasive  drilling  based on a pre-operative J-shape trajectory planned in CT scan of a patient followed by appropriate calibration,  registration, and navigation steps to safely execute the planned trajectory in real-time using our unique robotic setup. 
We holistically assess the performance of this framework on two different levels of vertebral phantoms  custom-designed based on CT scan of real patients.

\section{Patient-Specific Autonomous Robotic Framework}

Our patient-specific image-guided autonomous robotic framework is shown  in Fig. \ref{fig:framework}. As illustrated, the framework includes several modules including the planning, calibration and registration, and robotic drilling modules that are working in tandem to ensure an autonomous and accurate spinal fixation procedure. The envisioned robotic drilling  process starts with obtaining a quantitative CT (QCT) scan of a patient to obtain the spatial map of vertebral bone density \cite{screw_place,Shah2011ImagingOS} and planning a desired J-shape drilling trajectory in the CT image. Of note, the considered desired trajectory have a J-shape (i.e., a straight piece followed by a curved shape) because of the anatomy of vertebra and particularly the straight geometry of pedicle canal (see Fig. \ref{fig:plots}). The exact design of the J-shape trajectory   can be directly determined by the surgeon considering the spatial map of bone density   or after performing a biomechanical analysis on a quantitative/CT scan of the patient \cite{screw_place,Shah2011ImagingOS}. In this study, we are assuming a desired trajectory is obtained through either of these methods. Next, we integrate a bespoke curve drilling robot with the robotic manipulator and then perform hardware calibration of this robotic system with the use of an optical navigation system. Finally, the information  (i.e., anatomical geometry) gathered from the patient vertebra using a digitizer are used to register the patient in real space with the pre-operative CT scans of the patient. Next, the robot is commanded to autonomously perform the drilling process. The following sections describe these steps and components of the proposed frameworks in detail.

 \subsection{Integrated Robotic Framework Utilizing CT-SDR}
The proposed drilling robot, shown in Fig. \ref{fig:set-up}, 
consists of a redundant robotic manipulator with a unique CT-SDR. The robotic manipulator is used to properly position the CT-SDR in the workspace and provide extra required degree-of-freedom  (DoF)  to properly approach the vertebra based on the planned trajectory. As illustrated in Fig. \ref{fig:set-up}~A~\&~B, the CT-SDR is made of pre-shaped curved nitinol tubes heat-treated specifically based on the patient's vertebral anatomy (obtained through QCT scan) and its complementary planned J-shape trajectory in the planning module. The curved tube is nested and constrained inside a straight stainless steel tube and is used to  steer/guide a custom-designed flexible drill bit. The nitinol tube (Euroflex GmbH, Germany) for this study had 3.61 mm outer diameter with a 0.25 mm wall thickness and 70
mm length. The stainless steel tube had 80 mm length, 3.175 mm outer diameter, and 1.25 mm wall thickness (89895K712, McMaster-Carr).
As shown in Fig. \ref{fig:set-up}~A~, the flexible drill bit is made of a flexible shaft (Asahi Intec. USA, Inc.) laser welded to a machined drill bit (42955A26, McMaster-Carr). Of note, the dimensions of the  tubes and flexible drill bit is determined based on the geometry of the target vertebra and planned J-shape trajectory \cite{Sharma_tbme_2022}. The operating principle of the CT-SDR is  simple yet very intuitive making its use very easy. As shown in Fig. \ref{fig:set-up}B, the initially constrained curved nitinol tube is pushed out using a NEMA 17 stepper motor with a ball screw and a linear rail (B085TG12D1, Amazon) while steering the flexible rotating drill inside the hard tissue, therefore, precisely creating the curved piece of the J-shape trajectory. 
The straight piece of this trajectory is created when the nitinol tube is completely constrained in the stainless steel and the robotic manipulator inserts the CT-SDR through the pedicle canal.

\subsection{Hardware Calibration}

\begin{figure}[t!] 
	\centering 
	\includegraphics[width=1.0\linewidth]{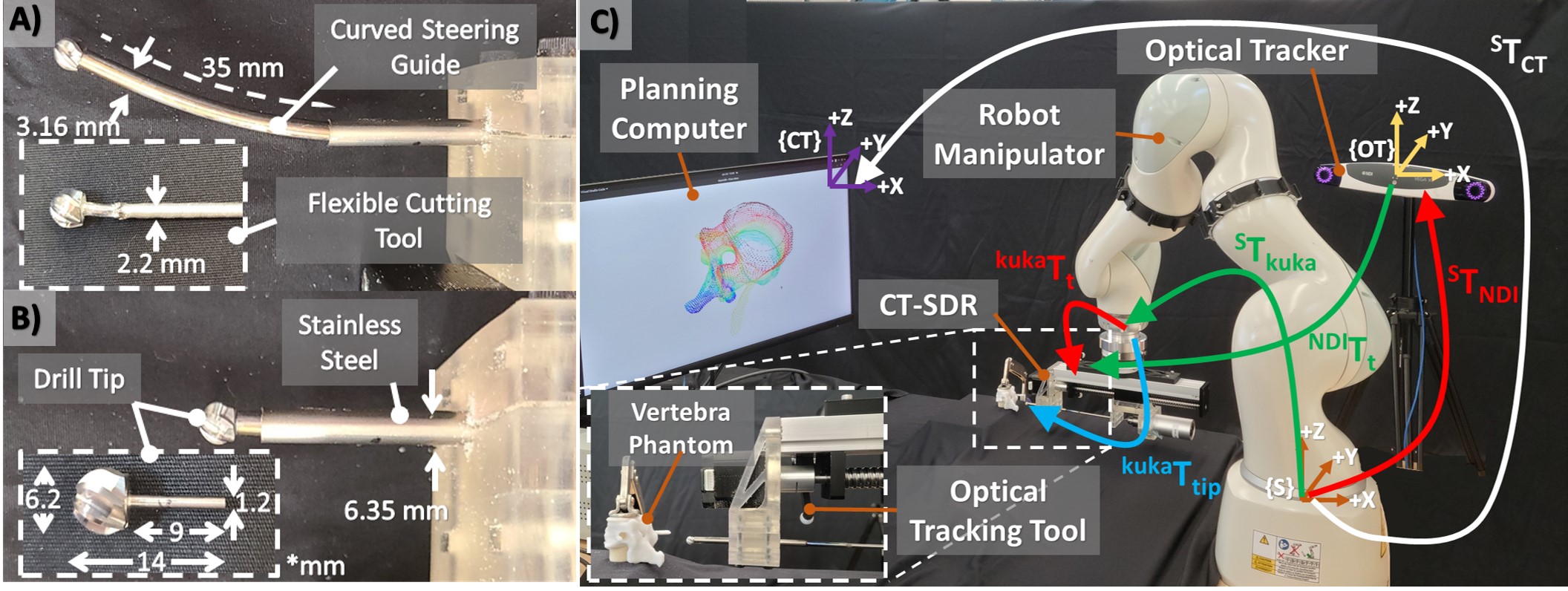}
	\caption{The experimental setup used for performing the experiments: (A\&B) Curved motion by the CT-SDR tip as it is actuated during drilling, with detailed view of the flexible cutting tool. (C) A view of the experimental set-up with the required known and calculated transformations. Known transformations are shown in green, blue transformations come from the performed pivot calibration, red from the performed hand-eye calibration, and white from the performed ICP registration. 
 }
	\label{fig:set-up}
\end{figure}

\textbf{Pivot Calibration:} Using an optical tracker  $\{OT\}$ (shown in Fig. \ref{fig:set-up}) and through the process of  pivot calibration formulated in  equation (\ref{eq:pivot_eq}), we were able to determine the positioning of the CT-SDR's end-effector $x_{tip}\in\mathbb{R}^3$ as related to the robotic manipulator's end effector. In this equation, $x_{pivot}\in\mathbb{R}^3$ indicates the pivot location with respect to the base of the robot $\{S\}$. For each collected pose, the transformation of the robot's end effector with respect to the base frame, $\{S\}$, is represented by its rotation $R_i\in SO(3)$ and translation $p_i\in\mathbb{R}^3$. Also, $I\in SO(3)$ represents the Identity matrix.
In Fig. \ref{fig:set-up} this transformation (${}^{kuka}T_{tip}\in SE(3)$) is marked with a blue arrow. 

\textbf{Hand-Eye Calibration:} A hand-eye calibration was also performed to identify the transformation between the optical tracking system    and the robotic manipulator \cite{Open3D}. This process was carried out using the $AX=ZB$ calibration in (\ref{eq:axzb}). In our formulation and as shown in Fig. \ref{fig:set-up}, ${}^{t}T_{NDI}\in SE(3)$ represents the sets of transformations between the optical tracker and the tracking tool attached to the CT-SDR, and ${}^{kuka}T_{S}\in SE(3)$ is the  transformation between the robot's end effector and the base frame $\{S\}$.
In Fig. \ref{fig:set-up}, the  calculated/unknown transformations $X$ and $Z$ (${}^{S}T_{NDI}\in SE(3)$ and ${}^{kuka}T_{t}\in SE(3)$) are marked with red arrows.

\begin{minipage}{.45\linewidth}
\begin{equation}\label{eq:pivot_eq}
\begin{bmatrix} ... & ... \\ R_i & -I \\ ... & ... \end{bmatrix} \begin{bmatrix} x_{tip} \\ x_{pivot} \end{bmatrix} = \begin{bmatrix} ... \\ -p_i \\ ... \end{bmatrix}
\end{equation}
\end{minipage}
\begin{minipage}{.45\linewidth}
\centering
\begin{equation}\label{eq:axzb}
\begin{bmatrix} ... \\ {}^{t}T_{NDI} \\ ... \end{bmatrix} 
\begin{bmatrix} X \end{bmatrix} =
\begin{bmatrix} Z \end{bmatrix} 
\begin{bmatrix} ... \\ {}^{kuka}T_{s} \\ ... \end{bmatrix}
\end{equation}
\end{minipage}

\subsection{Patient-CT Registration Using a Dual-Stage ICP}
To execute the planned J-shape trajectory in the patient's QCT scan using the proposed robotic system, we first need to register the patient's vertebra to the planned J-shaped trajectory. To this end, we need to calculate the transformation ${}^{S}T_{CT}\in SE(3)$  indicated in Fig. \ref{fig:set-up}. 
The process starts with using a calibrated digitizer tool to collect a set of points across the surface of the vertebra. These points were transformed from the optical tracker frame $\{OT\}$ to the world frame $\{S\}$ to ensure our calculated transformations would move between the  scanned frame $\{CT\}$ and world frame $\{S\}$ (see Fig. \ref{fig:set-up}). 
After collecting the data using digitizer, we performed a dual-stage registration process using the Iterative Closest Point (ICP) algorithm \cite{ICP_ZHOU202263}  to align the obtained partial set of points with the patient's CT scan data.

ICP algorithm aligns a subset of data points with a reference data set while attempting to minimize Euclidean distances between matching point pairs.
However, one of the limitations to the accuracy of ICP algorithms is that they can produce results with large error if the transformation required between the two data sets is particularly large \cite{ICP_ZHOU202263}. To address this issue, we first performed a loose global registration between the two sets of points. To this end, three points from the digitized scan and three corresponding points in the digital render were selected to create an initial alignment between the two data sets. This initial alignment formed the initial transformation matrix used to create the first correspondence set from the digital point cloud $P$ and the transformed digitized point cloud $Q$, using $\kappa=\{(p,q)\}$, where $p\in \mathbb{R}^3$ and $q\in \mathbb{R}^3$ represent points in the  data sets $P$ and $Q$, respectively. ICP algorithms then iteratively update the calculated transformation by minimizing the utilized objective function over the correspondence set. For the ICP algorithm in this paper, we utilized the objective function $E(T)=\sum_{(p,q)\in\kappa} ||p-Tq||^2$.
The resulting transformation matrix calculated from the ICP algorithm is shown as a white arrow in Fig. \ref{fig:set-up}.

\begin{figure}[t!] 
	\centering 
	\includegraphics[width=1.0\linewidth]{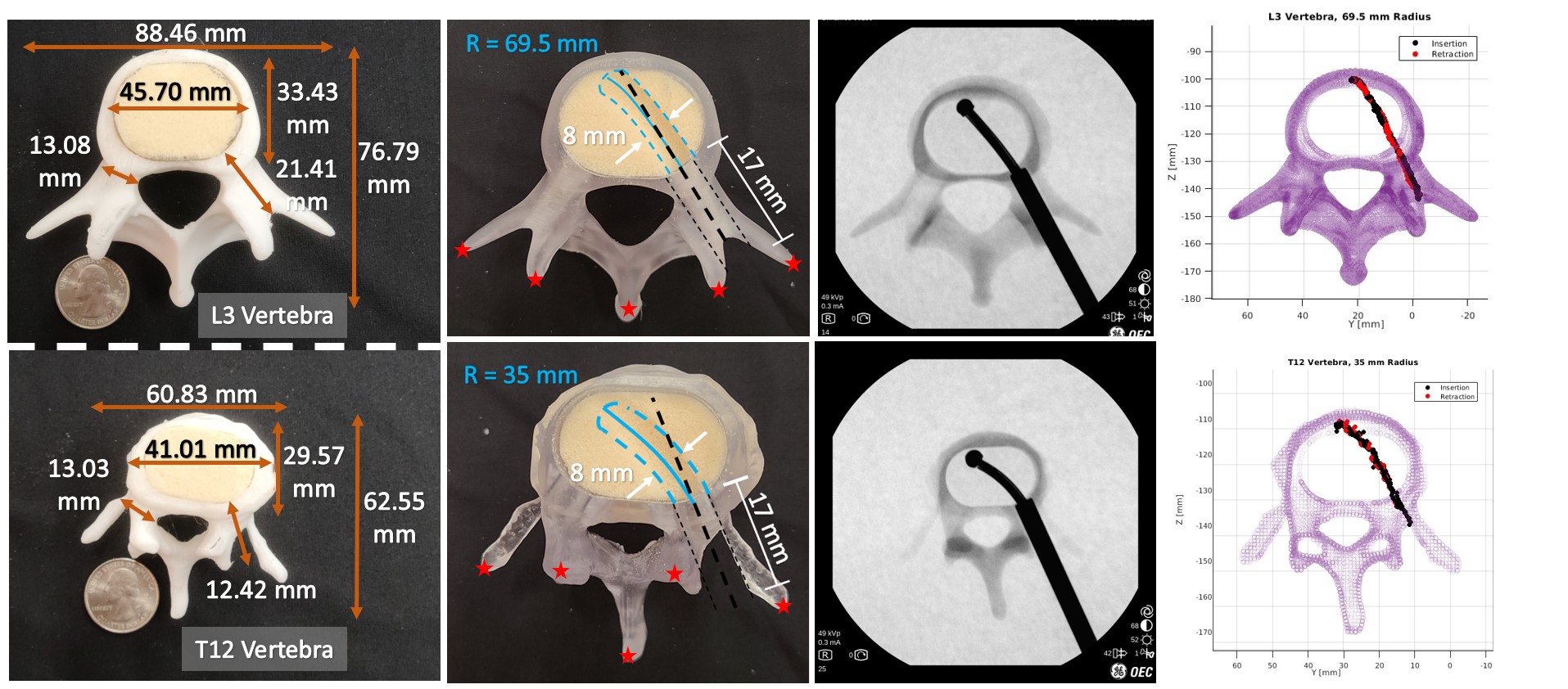}
	\caption{The test specimen used during testing along with an ideal cross-section view, x-ray view of the drilled trajectory, and plotted trajectories from the magnetic tracker. The 5 hardware evaluation points are indicated by red  $\star$.}
	\label{fig:plots}
\end{figure}
\section{Evaluation Experiments}
\subsection{Experimental Setup}

Fig. \ref{fig:set-up} shows the experimental setup used for performing the experiments. The CT-SDR was affixed to the end effector of a KUKA LBR med  7 DoF robotic arm. For our experiments, following the procedure described in \cite{hodgson2001fabrication}, we heat treated  two pre-curved nitinol tubes with 35 mm and 69.5 mm radius of curvature and 35 mm in length.
The  optical tracking system (NDI Polaris Vega, Northern Digital Inc.) provided continuous location tracking of the rigid body attached  to the CT-SDR in the workspace. The points for vertebral phantom registration were collected using the optical tracking digitizer tool. The vertebral phantoms shown in Fig. \ref{fig:plots} - representing L3 and T12 vertebrae - were custom-designed based on real CT scans of patients obtained using the protocol  approved by the review board of the university (IRB ID:STUDY00000519).  As shown, the cortical layer of the vertebra was 3D-printed using the PLA material and Raise3d printer (Raise 3D Technologies, Inc.), leaving a  space for placing simulated cancellous bone of the vertebral body made by PCF 5 Sawbone bio-mechanical bone phantom (Pacific Research Laboratories, USA).  Notably, PCF 5 was selected as it simulates a bone with  osteoporosis \cite{ccetin2021experimental}. Pilot hole channels served as entry points for the J-shape trajectory's straight section.
Several points of interest on these phantoms were identified with respect to the CT scan frame $\{CT\}$, including the center of the entry hole, and 5 locations of posterior elements of each vertebra. These points, marked by $\star$  in Fig. \ref{fig:plots}, were used to calculate the accuracy of our calibration and registration steps prior to experimentation.

\subsection{Experimental Procedure}
\subsubsection{Calibration and Registration Verification:}
After performing the calibration and registration steps, we verified its accuracy by aligning the vertebrae phantom's surface points, obtained via digitizer and optical tracking, with a  scan of the vertebra. Using established transformations, the digitizer was used to indicate the 5 points on the vertebra phantom, maintaining their integrity in the optical tracking reference frame. These points commanded the robotic system to position the CT-SDR accordingly. Comparing the actual CT-SDR tip positions with the ground truth digitized locations, we calculated error as an averaged Euclidean distance, providing a reliable assessment of our system's accuracy through calibrated transformations summarized in Table \ref{tab:data}.

\begin{table}[t] 
    \begin{center}
   \setlength{\tabcolsep}{1.2pt}%
    \caption{Average error across all performed tests on both vertebral phantoms in mm.}
    \label{tab:data}
    \begin{tabular}{c|c|c|c|c|c}
        Vertebra&ICP&Total Calibration&Ideal&Actual&Procedure\\
        Level&Error&Error&Trajectory&Trajectory&Time\\
        \hline
        L3&0.73$\pm$0.07&4.04$\pm$0.38&69.5&69.39$\pm$2.03&69 s\\
        T12&0.72$\pm$0.02&2.94$\pm$0.29&35&28.38$\pm$4.72&69 s\\
        \hline
    \end{tabular}
    \end{center}
\end{table}
\subsubsection{Autonomous Drilling Performance Evaluation:}
After obtaining accuracy of the calibration and registration steps, each vertebra was drilled using the robotic system. In the performed experiments, the robotic manipulator first aligned the CT-SDR's tip with the indicated entry point from the 3D scan, while considering a 10 mm distance  to prevent the system colliding with the phantom prior to the drilling. 
Upon reaching the target location, the CT-SDR's drill started drilling the phantom with approximately 8250 rpm. The  robotic manipulator  began the creation of the J-shape trajectory by inserting the CT-SDR's tip straight into the vertebra phantom's pedicle at 1 mm/s for 27 mm (17 mm + 10 mm buffer). This straight trajectory used the CT-SDR's outer stainless steel tube as the main guide. Next, the manipulator was held  steady and the CT-SDR made the 35 mm long curved section of the J-shape trajectory by inserting the  nitinol tube through the vertebral body at a speed of 2.5 mm/s. The curved trajectory utilized a 69.5 mm radius curved guide for the L3 vertebra specimen and a 35 mm curved guide for the T12 vertebra. we repeated each test 3 times.
The drilled trajectory's accuracy was assessed by passing a magnetic tracker (NDI Aurora, Northern Digital Inc.) through the drilled path, and the collected points were measured for accuracy evaluation. An ICP registration was also performed on the test specimen between the magnetic tracker and the phantom's CT scan to create the plots of the drilled trajectories shown in Fig. \ref{fig:plots}. Table \ref{tab:data} summarizes the obtained errors  as well as the procedure time. Fig. \ref{fig:plots} also shows the CT-SDR's drilling trajectory   obtained by C-arm X-ray machine (OEC One
CFD, GE Healthcare) as well as the cross section of the drilled trajectories inside the Sawbone phantoms.

\section{Results and Discussion}

 Each performed calibration contributed toward the overall performance of the drilling system. As shown in Table \ref{tab:data},  we obtained about 4.04$\pm$0.38 mm  and 2.94$\pm$0.29  mm overall calibration and registration error  for the L3 and T12 vertebrae, respectively. 
The root mean square errors (RMSE) of the correspondence sets after performing three trials of ICP registration were less than 1 mm, with the lowest error at 0.642 mm and the maximum at 0.819 mm, both seen on the L3 vertebra. This error is similar to that seen in other registration algorithms, such as \cite{Sefati_TRO}. The verification steps performed  based on the considered 5 points (indicated by red  $\star$ in Fig. \ref{fig:plots}) demonstrated an average error of 3.49$\pm$0.64 mm. 
Given the 8 mm diameter of the drilled tunnel (refer to Fig. \ref{fig:plots}), the pedicle canal of the vertebrae measuring approximately 13 mm (refer to Fig. \ref{fig:plots}), and the overall verification error of 3.49$\pm$0.64 mm, it is assured that there is a sufficient safety margin. This margin aligns with the consensus among surgeons that a medial pedicle perforation under 4 mm does not jeopardize the adjacent neural structures \cite{Gelalis2012AccuracyOP}. 
The measured drilling trajectories using the magnetic tracker (shown in Fig. \ref{fig:plots})  demonstrate that the trajectories drilled into the L3 vertebra has a better accuracy compared with  the T12 vertebra, with an error of 0.11 mm in the created radius of curvature. We  believe the main source of the higher error for the 35 mm trajectory is coming from the magnetic tracking system as the complementary  X-ray  and cross section views of the drilling trajectory (shown in Fig. \ref{fig:plots}) clearly illustrating the curvature of the drilled tunnel. Moreover, the procedure time for each vertebra, calculated from the time to enter the vertebra, drill the curved trajectory, and fully retract the system, was identical in both cases as the curved distance was the same for both vertebrae. The system's insertion time was approximately 35 s, significantly faster than the 300$+$ s drilling times observed in \cite{alambeigi2017curved} for comparable drilled lengths using tendon-driven manipulators. This clearly demonstrates the out-performance of the proposed autonomous drilling system compared with the similar systems in the literature.

\section{Conclusion and Future Work}

In this study, we proposed an image-guided autonomous and patient-specific framework to perform an unprecedented PSF procedure in osteoporotic patients  using a robotic system featuring a CT-SDR. By performing appropriate registration and calibration steps, we achieved average total calibration error   as low as 3 mm  for drilling unique J-shape trajectories in a T12  veretebra  with a drilling time  less than 70 s. In the future, we plan to improve the obtained drilling accuracy and perform a realistic cadaveric and animal studies using our proposed robotic framework.  We will also develop an end-to-end system by including a biomechanics-aware planning module to the proposed framework.

\bibliographystyle{splncs04}
\bibliography{main}

\end{document}